%% file: Formatting-Instructions-LaTeX-2021.tex
\def\year{2021}\relax
\pdfoutput=1
\documentclass[letterpaper]{article} 
\usepackage{aaai21}  
\usepackage{times}  
\usepackage{helvet} 
\usepackage{courier}  
\usepackage{amsmath}
\usepackage[hyphens]{url}  
\usepackage{graphicx} 
\usepackage{natbib}  
\usepackage{caption} 

\usepackage[utf8]{inputenc} 
\usepackage[T1]{fontenc}    
\usepackage{hyperref}       
\usepackage{url}            
\usepackage{booktabs}       
\usepackage{amsfonts}       
\usepackage{nicefrac}       
\usepackage{microtype}      

\usepackage{hyperref}
\usepackage{url}
\usepackage{algorithm}
\usepackage{algorithmic}
\usepackage{amsmath}
\usepackage{booktabs}       
\usepackage{multirow}
\usepackage{microtype}
\usepackage{graphicx}
\usepackage{subfigure}
\usepackage{booktabs} 
\usepackage[dvipsnames]{xcolor}
\usepackage{hyperref}
\usepackage{amsmath}
\usepackage{float}
\usepackage{caption}

\DeclareMathOperator*{\argmin}{arg\,min}

\usepackage{amssymb}
\begin{document}
\title{Semi-Supervised Few-Shot Learning with Prototypical Random Walks }
\author{

    Ahmed Ayyad${}^{1}$,
    Yuchen Li${}^{3}$,
    Raden Muaz${}^{3}$\\ 
    Shadi Albarqouni${}^{1\ 2}$ \thanks{Shared senior authorship},
    Mohamed Elhoseiny${}^{3\ *}$

}
\affiliations{

    ${}^{1}$Technical University of Munich, Munich, Germany\\
    ${}^{2}$Helmholtz AI, Helmholtz Center Munich, Neuherberg, Germany  
    \\
      ${}^{3}$King Abdullah University of Science and Technology (KAUST), Thuwal, Saudi Arabia \\

    \texttt{a.3ayad@gmail.com, yuchen.li@kaust.edu.sa, raden.m.muaz@gmail.com} \\
    \texttt{shadi.albarqouni@helmholtz-muenchen.de, mohamed.elhoseiny@kaust.edu.sa}

}

\maketitle

\begin{abstract}
Recent progress has shown that few-shot learning can be improved with access to unlabelled data,  known as semi-supervised few-shot learning(SS-FSL). We introduce an SS-FSL approach, dubbed as Prototypical Random Walk Networks(PRWN), built on top of Prototypical Networks (PN). We develop a random walk semi-supervised loss that enables the network to learn representations that are compact and well-separated.
Our work is related to the very recent development on graph-based approaches for few-shot learning. However, we show that compact and well-separated class representations can be achieved by modeling our prototypical random walk notion without needing additional graph-NN parameters or requiring a transductive setting where a collective test set is provided. Our model outperforms baselines in most benchmarks with significant improvements in some cases. Our model, trained with 40$\%$ of the data as labeled, compares competitively against fully supervised prototypical networks, trained on 100$\%$ of the labels, even outperforming it in the 1-shot mini-Imagenet case with 50.89$\%$ to 49.4$\%$ accuracy. 
We also show that our loss is resistant to distractors, unlabeled data that does not belong to any of the training classes, and hence reflecting robustness to labeled/unlabeled class distribution mismatch.  Associated github page can be found at   \url{https://prototypical-random-walk.github.io}.
\end{abstract}

\section{Introduction}

Few-shot learning is an artificial learning skill of rapidly generalizing from limited supervisory data (few labeled examples), typically without the use of any unlabeled data \cite{koch2015siamese, miller2000learning, lake2011one}. Our work is at the intersection between few-shot learning and semi-supervised learning where we augment the capability of few-shot artificial learners with a learning signal derived from unlabeled data. 


\begin{figure*}
\vspace{-1.5mm}
    \centering
    \includegraphics[width=1.03\textwidth]{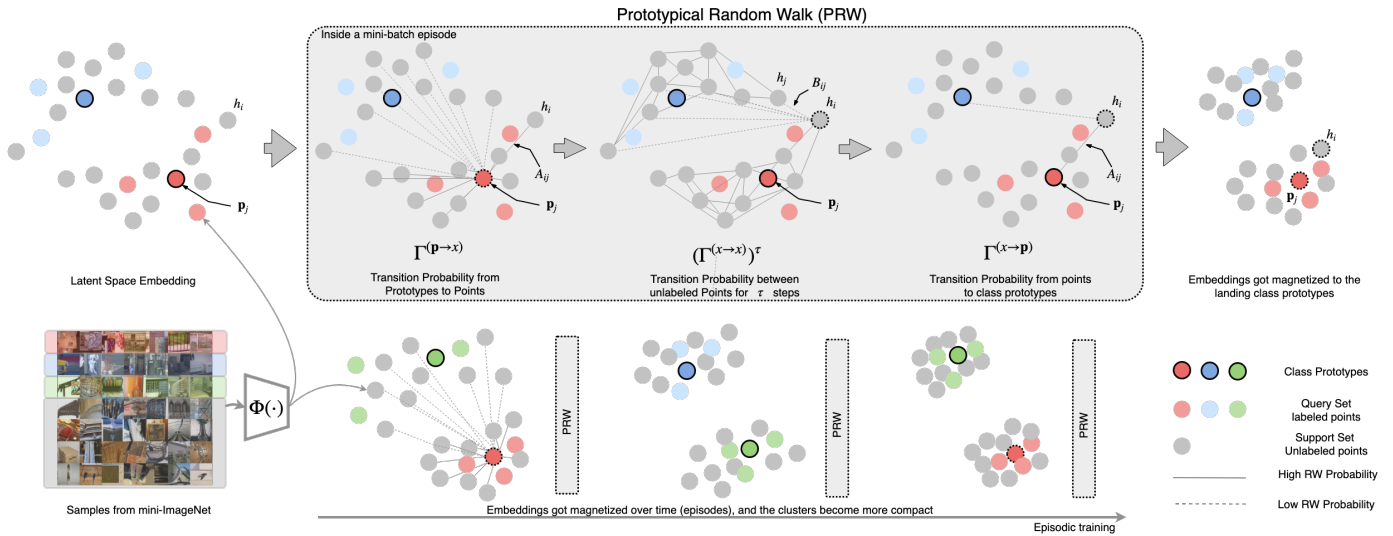}
    
    \caption{ Our PRW aims at maximizing the probability of a random walk begins at the class prototype $\mathbf{p}_j$, taking $\tau$ steps among the unlabeled data, before it lands to the same class prototype. This results in a  more discriminative representation, where the embedding of the unlabeled data of a particular class got magnetized to its corresponding class prototype, denoted as \emph{prototypical magnetization}.}
    \label{fig:teaser}
    \vspace{-1.5mm}
\end{figure*}

\emph{Semi-supervised Few-shot Learning (SS-FSL):} Few-shot learning methods typically adopt a supervised learning setup (e.g., ~\cite{vinyals2016matching,ravi2017optimization,snell2017prototypical}),  recently,~\cite{ren2018metalearning} and \cite{zhang2018metagan} developed Semi-supervised few-shot learning approaches that can leverage additional unlabeled data. The machinery of both approaches adopts a meta-learning episodic training procedure with integrated learning signals from unlabeled data. \cite{ren2018metalearning} build on top of prototypical networks(PN)~\cite{snell2017prototypical} so better class prototypes can be learned with the help of the unlabeled data. \cite{zhang2018metagan} proposed a GAN-based approach, Meta-GAN, that helps to make it easier for FSL models to learn better decision boundaries between different classes. 

In this work, we propose Prototypical Random Walks (PRW),  as an effective graph-based learning signal derived from unlabeled data. 
Our approach improves few-shot learning models by a prototypical random walk through the embeddings of unlabeled data starting from each class prototype passing through unlabeled data in the embedding space and encourages returning to the same prototype at the end of the prototypical walk (\emph{cf.}~Fig.~\ref{fig:teaser}). 
This PRW learning signal promotes a latent space where points of the same class are compactly clustered around their prototype while being well isolated from other prototypes. We sometimes refer to this discriminative attraction to class prototypes as \emph{prototypical magnetization}.

Since the PRW loss is computed over a similarity graph involving all the prototypes and unlabeled points in the episode, it takes a global view of the data manifold. Due to the promoted \emph{prototypical magnetization} property, this global view enables more efficient learning of discriminative embeddings from few examples, which is the key challenge in few-shot learning. In contrast, there are local SSL losses, where the loss is defined over each point individually, most notable of those approaches is the state-of-the-art Virtual Adversarial Training (VAT)~\cite{miyato2018virtual}.
We show that in the FSL setting, our global consistency guided by our prototypical random walk loss adds a learning value compared to local consistency losses as in VAT~\cite{miyato2018virtual}.

\textbf{Contribution.} We propose Prototypical Random Walk Networks (PRWN) where we promote \emph{prototypical magnetization} of the learning representation. We demonstrate the effectiveness of PRWN on popular few-shot image classification benchmarks. We also show that our model trained with a fraction of the labels is competitive with PN trained with all the labels. Moreover, we demonstrate that our loss is robust to "distractor" points which could accompany the unlabeled data yet not belong to any of the training classes of the episode.

\section{Approach}
\label{approach}

We build our approach on top of Prototypical Networks (PN)~\cite{snell2017prototypical} and augment it with a novel random walk loss leveraging the unlabeled data during the \textit{meta-training} phase.
The key message of our work is that more discriminative few-shot representations can be learned through training with prototypical random walks. 
We maximize the probability of a random walk which starts from a class of prototypes and walk through the embeddings of the unlabeled points to land back to the same prototype; see Fig.~\ref{fig:teaser}. 
Our random walk loss enforces the global consistency where the overall structure of the manifold is considered. 
In this section, we detail the problem definition and our loss.

\subsection{Problem Set-up}\label{PNapproachsub}
The few-shot learning problem may be formulated as training over distribution of classification tasks $\mathcal{P}_{train}(\mathcal{T})$, in order to generalize to a related distribution of tasks $\mathcal{P}_{test}(\mathcal{T})$ at test time.
 This setting entails two levels of {learning}; \emph{meta-training} is learning the shared model parameters(meta-parameters) to be used on future tasks, \emph{adaptation} is the learning done within each task. Meta-training can be seen as the outer training loop, while adaptation being the inner loop. 
 
Concretely, for $N_s$-shot $N_c$-way FSL, each task is an episode with a support set $\mathcal{S}$ containing $N_s$ labeled examples from each of $N_c$ classes, and a query set  $\mathcal{Q}$ of points to be classified into the $N_c$ episode classes. The support set is used for adaptation, then the query set is used to evaluate our performance on the task and compute a loss for meta-training. 

To run a standard FSL experiment, we split our datasets such that each class is present exclusively in one of our train/val/test splits. To generate a training episode, we sample $N_c$ training classes from the train split and sample $N_s$ samples from each class for the support set. Then we sample $N_q$ images from the same classes for the query set. Validation and test episodes are sampled analogously from their respective splits. 

Following the SS-FSL setup in~\cite{ren2018metalearning, zhang2018metagan,liu2019ICLR}, we split our training dataset into labeled/unlabeled; let $\mathcal{D}_{k,L}$ denote all labeled points $x \in class (k)$, and $\mathcal{D}_{k,U}$ be all unlabeled points $x \in class (k)$. Analogous notation holds for our support and query set, $\mathcal{S}$ and $\mathcal{Q}$. 
To set up a semi-supervised episode, we simply need to add some unlabeled data to the support set. For every class $c$ sampled for the episode, we sample $N_u$ samples from $\mathcal{D}_{c,U}$ and add them to $\mathcal{S}$. 
 In order to make the setting more realistic and challenging, we also test our model with the addition of \emph{distractor} data. Those are unlabeled points added to the support set, but not belonging to the episode classes. We simply sample $N_d$ additional classes, and sample $N_u$ points from each class to add to the support set. 
We present pseudo-code for episode construction in the supp. materials.


It is worth mentioning that the unlabeled data may be present at either train or test time, or both. At training time, we want to use the unlabeled data for meta-training i.e. learning better model parameters. For unlabeled data at test time, we want to use it for better adaptation, i.e. performing better classification on the episode's query set. Our loss operates on the meta-training level, to leverage unlabeled data for learning better meta-parameters. However, we also present a version of our model capable of using unlabeled data for adaptation, by using the semi-supervised inference from~\cite{ren2018metalearning} with our trained models.

\textbf{Prototypical Networks.} Prototypical networks~\cite{snell2017prototypical} aim to train a neural network as an embedding function mapping from input space to a latent space where points of the same class tend to cluster. The embedding function $\Phi(\cdot)$ is used to compute a \emph{prototype} for each class, by averaging the embeddings of all points in the support belonging to that class, $\mathbf{p}_c = \frac{1}{|\mathcal{S}_{c, L}|} \sum_{x_i \in \mathcal{S}_{c,L}}{\Phi(x_i; \theta)} $,
    
where $\mathbf{p}_c$ is the prototype for our $c$-th class, and $\theta$ represents our meta-parameters. Once prototypes of all classes are obtained, query points are also embedded to the same space, and then classified based on their distances to the prototypes, via a softmax function. For a point $x_i$, with an embedding $h_i = \Phi(x_i; \theta)$, the probability of belonging to class $c$ is computed by 
\begin{equation} \label{PNsoftmax}
\small
\begin{split}
    {z}_{i,c} =  p(y_c|x_i) =& \frac{\exp{(-d(h_i, \mathbf{p}_c))}}{\sum_{j= 1}^{N_c} \exp{(-d(h_i, \mathbf{p}_j)) }},  \\   \Tilde{\mathbf{p}}_{c} = & \frac{\sum_{x_i \in  {S}_{U}\cup{S}_{L}  } {h_i} \cdot {z}_{i,c}}{ \sum_{i=1}^{N} {z}_{i,c}}.
\end{split}
\end{equation}
where $d(\cdot,\cdot)$ is the Euclidean distance.
In the semi-supervised variant~\cite{ren2018metalearning}, PN use the unlabeled data to \emph{refine} the class prototypes. This is achieved via a soft K-means step. First, the class probabilities for the unlabeled data $z_{i,c}$ are computed as in Eq.\ref{PNsoftmax}, and the labeled points have a hard assignment, i.e. ${z}_{i,c}$ is 1 if $x_i \in class(c)$ and 0 otherwise. Then the updated prototype $\Tilde{\mathbf{p}}_{c}$ is computed as the weighted average of the points assigned to it; see Eq.~\ref{PNsoftmax}.
We can see this as a task \textit{adaptation} step, which does not directly propagate any learning signal from the unlabeled points to our model parameters $\theta$. In fact, it might be used only at the inference time, and results from~\cite{ren2018metalearning} show that it provides a significant improvement when used as such. When used during \emph{meta-training} by updating the model parameters from the unlabeled data, the performance improves only marginally (i.e., from 49.98\% to 50.09\% on mini-imagenet~\cite{vinyals2016matching}). 
While this approach is powerful as the \emph{adaptation} step, it fails to fully exploit the unlabeled data during \emph{meta-training}. 
\emph{SS-FSL with adaption at test time.} 
Our approach also allows using the former K-means refinement step at inference time, analogous to the 'Semi-supervised inference' model from \cite{ren2018metalearning}.  Orthogonal to~\cite{ren2018metalearning}, our approach can be thought of a meta-training regularizer that brings discriminative global characteristics to the learning representation efficiently . 

\subsection{Prototypical Random Walk}
\label{RW}

Given the class prototypes $\mathbf{p}_c$, computed using the labeled data in the support set $\mathcal{S}_{L}$, and the embeddings $h_i$ of unlabeled support set $\mathcal{S}_{U}$, we construct a similarity graph between the unlabeled points' embeddings and the prototypes. 
Our goal is to have points of the same class form a compact cluster in latent space, well separated from other classes. 
Our Prototypical Random Walk(PRW) loss aims to aid this by compactly attracting the unlabeled embeddings around the class prototypes, promoting well-separation(\emph{cf.}~Fig.~\ref{fig:teaser}). 

This notion is translated into the idea that a random walker over the similarity graph rarely crosses class decision boundaries. Here, we do not know the labels for our points or the right decision boundaries, so we cannot optimize for this directly. We basically imagine our walker starting at a prototype, taking a step to an unlabeled point, and then stepping back to a prototype. The objective is to increase the probability that the walker returns to the same prototype it started from; we will refer to this probability as the \textit{landing probability}. Additionally, we let our walker taking some steps between the unlabeled points, before taking a step back to a prototype.  


Concretely, for an episode with $N$ class prototypes, and $M$ unlabeled points overall, let $A \in \mathbb{R}^{M\times N}$ be the similarity matrix, such that each row contains the negative Euclidean distances between the embedding of an unlabeled point an d the class prototypes. Similarly, we compute the similarity matrix between the unlabeled points $B \in \mathbb{R}^{M\times M}$.  Mathematically speaking, $A_{i,j} = -\|h_i - \mathbf{p}_j\|^2,  B_{i,j} = -\|h_i - h_j\|^2$
where $h_i = \Phi(x_i) $ is the embedding of the $i$-th unlabeled sample, and $\mathbf{p}_j$ is the $j$-th class prototype.
The diagonal entries $B_{i,i}$ are set to a small enough number to avoid self-loop.    



Transition probability matrices for our random walker are calculated by taking a softmax over the rows of similarity matrices. For instance, the transition matrix from prototypes to points is obtained by softmaxing $A^T$,
$\Gamma^{(\mathbf{p} \rightarrow x)} = \textup{softmax}(A^T)$, 
such that $p(x_i | \mathbf{p}_j) = \Gamma^{(\mathbf{p} \rightarrow x)}_{j,i}$.
Similarly, transition from points to prototypes $\Gamma^{(x \rightarrow \mathbf{p})}$, and transitions between points $\Gamma^{(x \rightarrow x)}$, are computed by softmaxing $A$, and $B$, respectively. 
Now, we define our random walker matrix as
\begin{equation}\label{eq:rw}
\small
    T^{(\tau)} = \Gamma^{(\mathbf{p} \rightarrow x)} \cdot (\Gamma^{(x \rightarrow x)})^\tau \cdot \Gamma^{(x \rightarrow \mathbf{p})},
\end{equation}
where $\tau$ denotes the number of steps taken between the unlabeled points, before stepping back to a prototype. An entry $T_{i,j}$ denotes the probability of ending a walk at prototype $j$ \emph{given} that we have started at prototype $i$, and the $j$-th row is the probability distribution over ending prototypes, given that we started at prototype $j$.  
The diagonal entries of $T$ denote the probabilities of returning to the starting prototype; our landing probabilities. Our goal is to maximize those by minimizing a cross-entropy loss between the identity matrix $I$ and our random walker matrix $T$, dubbed as $L_{walker}$\footnote{To be exact, this is the average cross-entropy between the individual rows of $I$ and $T$, since those are probability distributions.}
\begin{equation}
\label{eq:RW_loss}
\begin{split}
  \mathcal{L}_{walker} =& \sum_{i=0}^{\tau} {\alpha^{i} \cdot H(I, T^{(i)})},  \,\,\,\, \mathcal{L}_{visit} = H{(\mathcal{U}, P)},    \, \,\,\, \\ 
  {\mathcal{L}_{RW}  =} & \mathcal{L}_{walker} + \mathcal{L}_{visit}, 
  \end{split}
\end{equation}
where $H(I , T) = - \frac{1}{N_c}  \sum_{i=0}^{N_c}{\log{ T_{i,i}}}$, and $\alpha$ is an exponential decay hyperparameter.
However, one issue with $L_{walker}$ loss, is that we could end up visiting a small subset of the unlabeled points.  To remedy this problem, \cite{8099557} introduce a 'visit loss', pressuring the walker to visit a large set of  unlabeled points. Hence, we assume that our walker is equally likely to start at any prototype, then we compute the overall probability that each point would be visited when we step from prototypes to points $P = \frac{1}{N_c} \sum_{i=0}^{N_c}{\Gamma^{(\mathbf{p} \rightarrow x)}_i }$, where $\Gamma^{(\mathbf{p} \rightarrow x)}_i$ represents a column of the matrix. Then we add $\mathcal{L}_{visit}$  as the standard cross-entropy between this probability distribution and the uniform distribution $\mathcal{U}$. Hence, our final random walk loss is $\mathcal{L}_{RW}$ is the sum of  $ \mathcal{L}_{walker}$ and $\mathcal{L}_{visit}$; see Eq~\ref{eq:RW_loss}.    
   
%

\textbf{Overall Loss.}
To put it all together, our objective function can be written as 
   $ \argmin_{\theta} \mathcal{L}_S  + \lambda {\mathcal{L}_{RW}}$, 
where $\lambda$ is a regularization parameter. While gradient of $\mathcal{L}_S = -\sum_{i=0}^{Q_L} y_i \log z_{i,c}$ provides the supervised signal, the gradient of $\mathcal{L}_{RW}$ encourages the \emph{``prototypical magnetization''} property guided by our random walk.
This loss is minimized in expectation over randomly sampled semi-supervised episodes from our training data.



\section{Related Work}
\label{background}

\textbf{Associative learning and local consistency in Semi-Supervised Learning} 
SSL contains a rich toolbox of principles and techniques to leverage unlabeled data to learn better discriminative embeddings. The core idea is that similar inputs tend to close in embedding space, measured with metrics such as Euclidean distance and KL-divergence. Loss functions are designed to further encourage inputs belonging to same class to cluster together in embedding space, such as triplet loss, mean teacher and learning by association \cite{haeusser2017learning}.

Particularly, our work can be seen as generalization of learning by association \cite{haeusser2017learning}, which is special case of PRWN with walk step size of 1. By using walk step size larger than 1, we effective increase the local receptive field to further associate and cluster similar inputs together in embedding space

\textbf{Semi-Supervised Few-Shot Learning} 
In few-shot learning task, the model is assumed to be meta-learned (read: pretrained) to reach a good starting point, so that in testing phase, learning with very few samples possible. \cite{miyato2018virtual, Kamnitsas2018SemiSupervisedLV, 8099557}. 

However, few-shot learning assumes the dataset to be meta-learned is sufficiently large and fully-labeled.
Hence, ~\cite{ren2018metalearning} introduced the SS-FSL setting, so that FSL is feasible where labeled data is scarce.  This setting combines both SS and FSL settings, where the meta-learning phase leverage unlabeled data to better meta-learn the task.

SS-FSL is also applied in other ways such as self-teaching \cite{li2019learning} and adaptive subspace \cite{simon2020adaptive}, and applied in other tasks, such as transfer learning \cite{yu2020transmatch} and image translation \cite{wang2020semi}. Orthogonal to these   developments, our goal is  to show that learning representations can be efficiently improved by prototypical random walk loss.

\textbf{Application of Random Walk for Associative Learning} 
Our work is also similar to application of random walk for person re-identification \cite{shen2018deep}, where random walk is used to re-rank and find best match given input \textit{probe image} relative to collection of known \textit{gallery images}. However, our focus is on applying PRWN for semi-supervised few-shot learning task, that is to leverage unlabeled data for meta-training  the model to good initial parameters, so that it is able to learn with few samples during test time.

\section{Experiments on 2D synthetic datasets}
To gain an intuition on how the proposed method works, we performed experiments on 2D synthetic datasets to easily visualize how the decision boundary is formed.

The model is 3-layer MLP, 2-dimension input, 32-dimension hidden unit, 4-dimension output. It used negative Euclidean distance metric for its output. We used two datasets, and 3 models trained 300 epochs in each dataset: (1) Spiral dataset with 1000 points split to 7 labels (10\% labels + Random Walk, 10\% labels, and 100\% labels; 1 shot, 5 way, $\tau$ = 1);
 (2) Gaussian circle dataset 1000 points split to 3 labels. There were 3 models trained in each dataset (5\% labels + Random Walk, 5\% labels, and 5\% labels; 1 shot 3 way, $\tau$ = 1
) ;
\emph{cf.}~Fig. 2 shows the results, it can be seen that the proposed method can “connect the dots” of unlabeled points in green region and purple region, hence producing decision boundary similar to 100\% labels. In Gaussian circle dataset, random walk loss helps the model fits the circle more in just few epochs, but the model without random walk loss still has many 
mis-classified points and the circular outline is not obvious.

\begin{figure*}[htbp!]
    \centering
    \includegraphics[width=0.9\textwidth,height=9cm]{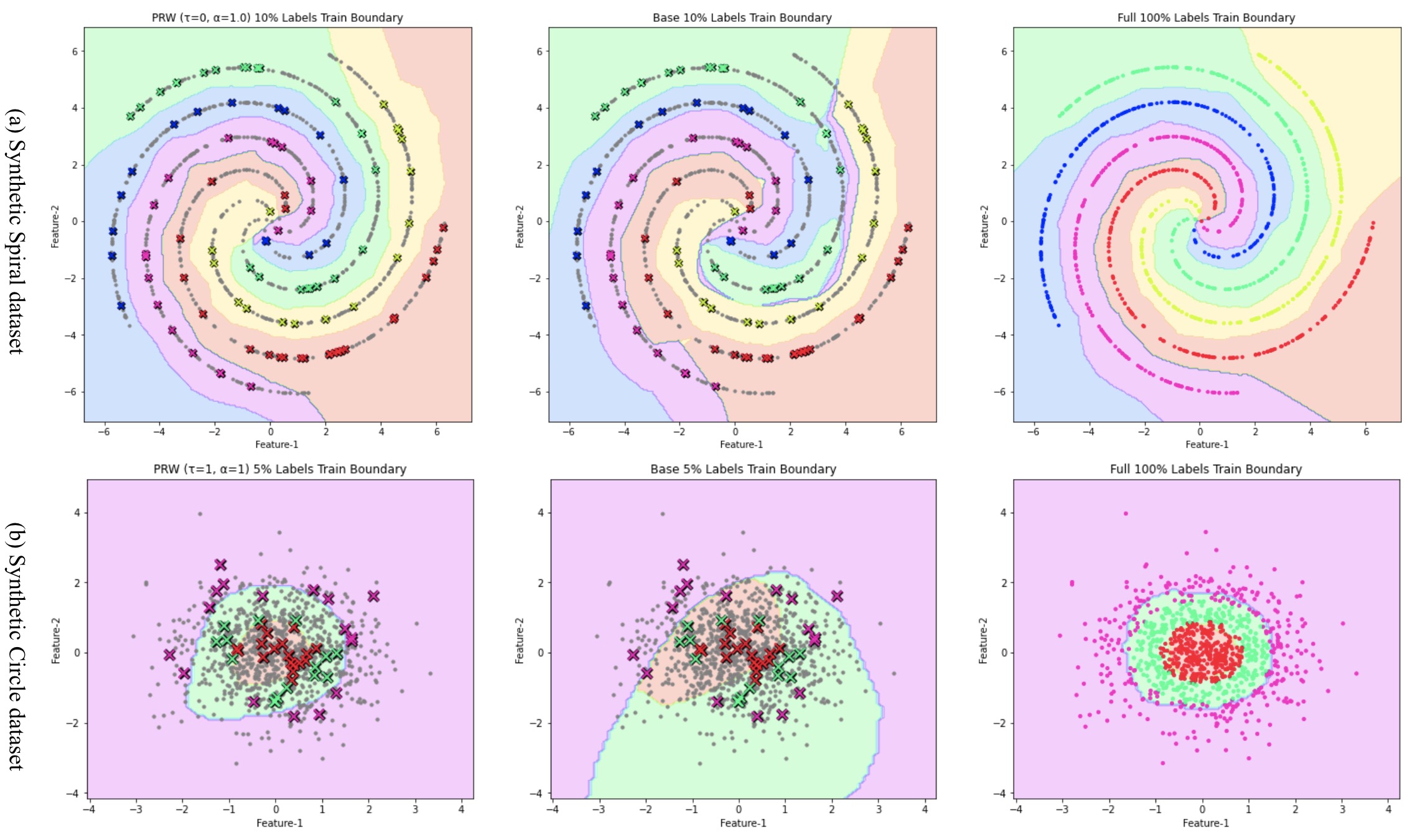}
    \caption{The colored points are labeled data and the grey points are unlabeled points. The toy experiments trains on spiral dataset (top row) and Gaussian circles dataset (bottom row). In spiral dataset, Left column: Prototypical Random Walk (PRW) 10\% labeled data; Middle column: 10\% labeled data; Right column: 100\% labeled data. In Gaussian circles dataset, Left column: Prototypical Random Walk (PRW) 5\% labeled data; Middle column: 5\% labeled data; Right column: 100\% labeled data}
    \label{fig:spiral}
\end{figure*}

\section{Experiments on Image Datasets}
\textbf{Overview.} In these experiments, we cover two main results: with and without distractors, where  distractors are present at train \emph{and} test time when applied. In each, we discuss experiments with and without \emph{semi-supervised adaptation} where additional unlabeled data are used at \textit{test} time. Note that whether or not unlabeled data is available at test time, we use the same trained model, the difference comes from adding the adaptation step  in Eq.~\ref{PNsoftmax} at test time to leverage that data.

\subsection{Experimental Setup}

\textbf{Datasets.} We evaluated our work on the two commonly used SS-FSL benchmarks {Omniglot}, {Mini-ImageNet}, and {tiered-ImageNet}. 
{Omniglot}~\cite{lake2011one} is a dataset of 1,623 handwritten characters from 50 alphabets. Each
character was drawn by 20 human subjects. We follow the few-shot setting proposed by~\cite{vinyals2016matching}, in which the images are resized to $28 \times 28$ px and rotations in multiples of $90^\circ$ are applied, yielding 6,492 classes in total. These are split into 4,112 training classes, 688 validation classes, and 1,692 testing classes.
{Mini-ImageNet} \cite{vinyals2016matching} is a modified version of the ILSVRC-12 dataset \cite{ILSVRC15}, in which 600 images, of size $84 \times 84$ px, for each of 100 classes were randomly chosen to be part of the dataset. We rely on the class split used by~\cite{ravi2016optimization}. These splits use 64 classes for training, 16 for validation, and 20 for test. 
{tiered-ImageNet}~\cite{ren2018metalearning} is also a subset of the ILSVRC-12 dataset \cite{ILSVRC15}. However, it is way bigger than the {Mini-ImageNet} dataset in the number of images; around 700K images, and the number of classes; around 608 classes coming from 34 high-level categories. Each
high-level category has about 10 to 20 classes and split into 20 training (351 classes), 6 validation (97 classes) and
8 test (160 classes) categories.

In our experiments, following ~\cite{ren2018metalearning,zhang2018metagan},  we sample 10\% and 40\%  of the points in each class to form the labeled split for Omniglot  and Mini-Imagenet, respectively; the rest forms the unlabeled split. 

\textbf{Implementation Details.} 
We have provided full details of our experimental setting including network architectures, hyperparameter tuning on the validation set in supp. materials.
For fair comparison, we opt for the same Conv-4 architecture~\cite{vinyals2016matching} appeared in the prior SS-FSL art~\cite{zhang2018metagan, ren2018metalearning}.

\textbf{Episode Composition.} All testing is performed on 5-way episodes for both datasets. Unless stated otherwise, the analysis performed in sections~\ref{semi-supervised meta results} \& \ref{sub:dist_analysis} are performed by averaging results over 300 5-shot 5-way mini-imagenet episodes from the \emph{test} split, with $N_u$=10. Further detail is in supp. materials.
All accuracies reported are averaged over 3000 5-way episodes and reported with 95\% confidence intervals. 

\textbf{Baselines.} We evaluate our approach on standard SS-FSL benchmarks and compare to prior art; PN~\cite{ren2018metalearning}, MetaGAN~\cite{zhang2018metagan}, and EGNN-Semi~\cite{kim2019edge}.  We also compare PRWN with 3 control models; the vanilla prototypical network (PN) trained on the fully labeled dataset, denoted PN$_{all}$ (the oracle), which is considered to be our \emph{target} model, a PN~\cite{ren2018metalearning} model trained only on the labeled split of the data (40\% of the labels), which is essentially PRWN without our random walk loss, and finally a PN trained with the state-of-the-art VAT~\cite{miyato2018virtual} and entropy minimization as a strong baseline; we denote it as PN$_{VAT}$. 

\begin{table*}[t!]
\centering
    \caption{Semi-Supervised Meta-Learning + Ablation Study} \label{tbl_SOA}
\resizebox{0.98\textwidth}{!}{
    \begin{tabular}{l c c c c c} 
     \toprule
     \multirow{2}{*}{Model}     & Omniglot &        \multicolumn{2}{ c }{Mini-Imagenet}  &  \multicolumn{2}{ c }{Tiered-Imagenet}\\ 
                                &  1-shot  &                 1-shot & 5-shot &                 1-shot & 5-shot\\
     \toprule
    PN$_{all}$\citep{snell2017prototypical}& 98.8                  & 49.4                  & 68.2 & 53.6 & 74.34\\
    \midrule
    PN~\citep{ren2018metalearning}   & 94.62 $\pm$ 0.09        & 43.61 $\pm$ 0.27        & 59.08 $\pm$ 0.22 &46.52$ \pm $ 0.52    & 66.15$ \pm $ 0.22\\
     \midrule    
    MetaGAN~\citep{zhang2018metagan}                          & 97.58 $\pm$ 0.07        & 50.35 $\pm$ 0.23        & 64.43 $\pm$ 0.27 & N/A & N/A\\
    EGNN-Semi~\citep{kim2019edge} &  N/A &  N/A & 62.52  $\pm$ N/A  & N/A & \textbf{70.98} $\pm$ N/A \\
    PN$_{VAT}$ (Ours)                     &  97.14 $\pm$ 0.16        & 49.18 $\pm$ 0.22            & 66.94 $\pm$ 0.20 & N/A & N/A\\ 
    PRWN (Ours)                      & \textbf{98.28} $\pm$ \textbf{0.15}     & \textbf{50.89} $\pm$ \textbf{0.22}   & \textbf{67.82} $\pm$ \textbf{0.19} & \textbf{54.87} $\pm$ \textbf{0.46}  & 70.52 $\pm$ 0.43\\
     \midrule
    PN + Semi-supervised inference\citep{ren2018metalearning}     & 97.45 $\pm$ 0.05  & 49.98 $\pm$ 0.34   & 63.77 $\pm$ 0.20 & 50.74 $\pm$ 0.75   & 69.37 $\pm$ 0.26\\
    PN + Soft K-means\citep{ren2018metalearning}                  & 97.25 $\pm$ 0.10 & 50.09 $\pm$ 0.45 & 64.59 $\pm$ 0.28 & 51.52 $\pm$ 0.36 & 70.25 $\pm$ 0.31\\
    PN + Soft K-means + cluster\citep{ren2018metalearning}        & 97.68 $\pm$ 0.07 & 49.03 $\pm$ 0.24 & 63.08 $\pm$ 0.18 & 51.85 $\pm$ 0.25 & 69.42 $\pm$ 0.17\\
    PN + Masked soft K-means\citep{ren2018metalearning}           & 97.52 $\pm$ 0.07 & 50.41 $\pm$ 0.24 & 64.39 $\pm$ 0.24 &  52.39 $\pm$ 0.44 & 69.88 $\pm$ 0.20\\
    TPN-Semi~\citep{liu2018learning} &  N/A &  52.78 $\pm$ 0.27 & 66.42 $\pm$ 0.21 &  55.74 $\pm$ 0.29 &  71.01 $\pm$ 0.23\\
    PRWN + Semi-supervised inference (Ours) & \textbf{99.23} $\pm$ \textbf{0.08}   & \textbf{56.65} $\pm$ \textbf{0.24}  & \textbf{69.65} $\pm$ \textbf{0.20} & \textbf{59.17} $\pm$ \textbf{0.41}  & \textbf{71.06} $\pm$ \textbf{0.39}\\
    \end{tabular}
}
\end{table*}


\subsection{Semi-supervised meta-learning without distractors}\label{semi-supervised meta results}

For experiments without semi-supervised adaptation, we observe from the third horizontal section of Table~\ref{tbl_SOA}, that PRWN improves on the previous state-of-the-art MetaGAN~\cite{zhang2018metagan}, and EGNN-Semi~\cite{kim2019edge} on all experiments, with a significant improvement on 5-shot mini-imagenet. It is worth mentioning that our PRWN has less than half the trainable parameters of MetaGAN which employs an additional larger generator. 

Experiments with semi-supervised adaptation are presented in the bottom section in Table~\ref{tbl_SOA}. Note that PRWN already improves on prior art without the adaptation. With the added semi-supervised adaptation, PRWN improves significantly, and the gap widens. On the 5-shot mini-imagenet task, PRWN achieves a relative improvement of 8,17\%, 4,86\%, and 8,28\% over the previous state-of-the-art, \cite{ren2018metalearning,liu2019ICLR,kim2019edge}, respectively. Similar behavior has been observed on tiered-ImageNet dataset outperforming existing methods in 1-shot classification and similar performance on 5-shot classification; note that standard deviation for \cite{kim2019edge} is not reported for 1 and 5-shot classification. 


\label{sub:analysis}
\begin{figure*}[ht]
\centering
\subfigure[Landing probabilities]{%
\includegraphics[width=0.32\textwidth]{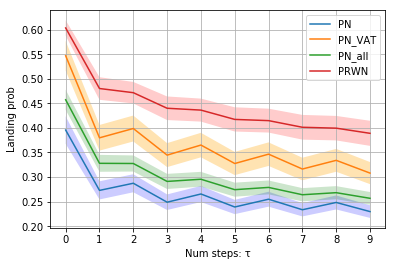}
}
\subfigure[Higher-Way Performance]{%
\includegraphics[width=0.32\textwidth]{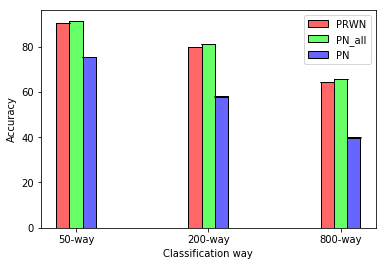}
}
\subfigure[Improvement over PN] {\includegraphics[width=0.32\textwidth]{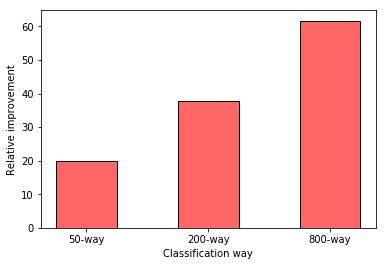}
}
\caption{\emph{(a)} Landing Probabilities on mini-ImageNet: The $x$-axis denotes the number of steps for the walk ($\tau$), and the $y$-axis shows the probability of returning to the right prototype. \emph{(b):} The Higher-Way performance on Omniglot as we increase the number of test classes $N_c$. \emph{(c):} The relative improvement of PRWN over PN as we increase the number of classes in Omniglot} 
\label{fig:landing}
\end{figure*}

{Ablation study.} From Table~\ref{tbl_SOA}, we can see that our PRW loss improves the baseline PN significantly, boosting the accuracy of PRWN up to {67.82\%} from 59.08\% on the 5-shot mini-imagenet for example. Moreover, while PN$_{VAT}$ proves a powerful model, competing with prior state-of-the-art, PRWN still beats it on all tests. Furthermore, We trained PRWN on mini-imagenet with only 20\% of the labels, and we obtain an accuracy of 64.8\% on the 5-shot task; outperforming the SOTA of 64.43\% which uses double the amount of labels. Most remarkably, PRWN performs competitively with the fully labeled PN$_{all}$, even {outperforming} it on 1-shot mini-imagenet.

\textbf{Local \& Global consistency Analysis.} To evaluate the global consistency,  we take a look into the behavior of our random walker for our various models. We compute the landing probability over the graphs they generate: the probability a random walker returns to the starting prototype, given by $Trace(T^{(\tau)})$ from Eq.~\ref{eq:rw}. We can see in 
Fig.~\ref{fig:landing} that even as $\tau$ grows, PRWN generates graphs with the highest landing probs. Following is PN$_{VAT}$, implying that enforcing local consistency still helps with global consistency.  We can also see that PN$_{all}$ also does better than PN, indicating that the addition of extra labeled data also improves global consistency. To evaluate local consistency and adversarial robustness of our various models, we compute their average VAT loss. Unsurprisingly, PN$_{VAT}$ performs best with {1.1} loss, following are both PRWN and PN$_{all}$ with {3.1 \& 2.91} respectively, then PN with {5.9}. We see again that improving global consistency helps with local consistency, and so does additional labeled data. 

\textbf{Discriminative Power. } In order to study our approach and baselines in a more challenging setup, we evaluate their performance on a Higher-Way classification. 
Fig.~\ref{fig:landing} shows that our model still performs better than the baseline and close to PN$_{all}$ (the oracle).  The accuracy of PRWN, PN$_{all}$, and PN, on 800-ways in Omniglot, are 64.43\%, 65.57\% and 39.84\%, respectively. 
In Fig.~\ref{fig:landing}, we show the relative improvement over PN reaching $\approx$ 60\% improvement on 800-ways classification. Similar behavior has been reported for mini-imagenet (See Supp. materials).
This shows the performance gain from our PRW loss is robust and reflects its discriminative power. 

\input{table2.tex}

\textbf{Transductive/Semi-supervised adaptation approaches.} Our approach is orthogonal and can be integrated with these methods~\citep{liu2019ICLR, ren2018metalearning, douze2018CVPR}. In fact, PRWN + semi-supervised inference is such an integration where K-means step is integrated from ~\citep{ren2018metalearning}. 
Tables \ref{tbl_SOA}, and \ref{tbl_dist} show that our network, combined with the K-means step at test time, perform far better than the networks trained with those adaptation methods.
This supports our hypothesis that semi-supervised adaptation like the K-means step fails to fully exploit the unlabeled data \textbf{during meta-training}.

\subsection{Semi-supervised meta-learning with distractors}
The introduction of distractors by~\citet{ren2018metalearning} was meant to make the whole setup more realistic and challenging. To recap, the distractors are unlabeled points added to your support set, but they do not belong to any of the classes in that set i.e. the classes you are currently classifying over. This "labeled/unlabeled class mismatch" was found by~\citet{realisticSSL} to be quite a challenge for SSL methods, sometimes even making the use of unlabeled data harmful for the model. 
We present our results in table \ref{tbl_dist}, where the top row is our model without test time adaptation, and we can see that it already beats the previous state-of-the-art below, which makes use of test time unlabeled data, even by a large margin in the 5-shot mini-imagenet with a relative improvement of 3,8\%, and 6,1\% on TPN-Semi~\cite{liu2019ICLR}, and PN+Soft K-Means~\cite{ren2018metalearning}, respectively. Moreover, it beats the MetaGAN~\cite{zhang2018metagan} model trained without distractors on all tasks, and in fact performs closely to our own PRWN trained without distractors (\emph{cf.} Table \ref{tbl_SOA}).

When we add the semi-supervised adaptation step, with distractors  present among unlabeled data at test time,  we see that our model does not benefit well from that step, and in the case of the 5-shot mini-imagenet, the performance is slightly harmed. 
Next subsection, we will explore why our model is robust to distractors during training, and how we can use the random walk dynamics to make the semi-supervised inference step useful when distractors are present.

\subsubsection*{Distractor analysis}
\label{sub:dist_analysis}
We hypothesize that the reason our PRWN is robust against distractors, is because our random walker learns to largely avoid distractor points, and as such they are not magnetized towards our class prototypes; if anything by learning to avoid them, the network is structuring the latent space such that points of each class are compact and well separated.  This comes as a by-product of the ``prototypical magnetization'' property that our  loss models. 

To test this hypothesis, we take a PRWN model trained with distractors, we sample test episodes including distractors ($N_d=N_c=5$), construct our similarity graph, and compute the probability that our random walker visits distractor versus non-distractor points. Concretely, we compute  $P = \frac{1}{N_c} \sum_{i=0}^{N_c}{\Gamma^{(\mathbf{p} \rightarrow x)} }$, where the summation is over the columns, and an entry $P_i$ represents the probability of visiting point $i$. We split $P$ into $P_{clean}$ and $P_{dist}$, containing the entries for non-distractor and distractor points. respectively. Both probabilities $p_{clean}$ and $p_{dist}$ should sum up to one. Whereas our baseline PN gets $p_{clean} = 0.67$, and PN$_{all}$ gets $p_{clean} = 0.76$, our PRWN gets $p_{clean} = 0.81$.  So we see our $\mathcal{L}_{RW}$ is not only an attractive force bringing points closer to prototypes, but it also has a repelling force driving irrelevant points away from prototypes. Note this is not only a feature of the network, it is a property of the loss function. For instance, the semi-supervised inference step \cite{ren2018metalearning} involves all points, distractor or not, equally in the prototype update, regardless of the geometry of the embeddings. 

\textbf{Distractors at semi-supervised inference.} 
We also performed  an experiment to further improve PRWN with the semi-supervised inference step. We exploit our random walk dynamics to order to filter out distractors. We compute the probability that a point is part of a successful walk; a walk which starts and ends at the same prototype. This is given by 
$S = \sum_{i=0}^{N_c} { \Gamma^{(\mathbf{p} \rightarrow x)} \odot   \Gamma^{(x \rightarrow \mathbf{p})} }$, where $\odot$ is the Hadamard product, and the summation is over the columns of the resulting matrix. Then we simply discard the points that scored below the median. With this little step, we see our PRWN + semi-supervised inference, become more robust to test time distractors, with {99.04\%} accuracy on omniglot,  {54.51\% \& 68.77\%} on mini-imagenet, and {57.97\% \& 69.74\%} on tiered-imagenet 1\&5-shot, respectively. This simple filtering step just improved on the distractor state-of-the-art as shown in  Table~\ref{tbl_dist} (last row). Note that our approach also outperform~\cite{liu2019ICLR} by a significant margin in 1-shot classification in all datasets and 5-shot classification Mini-Imagenet, while achieving similar performance on 5-shot Tiered-Imagenet.   

\section{Conclusion}

SS-FSL is a relatively unexplored yet challenging and important task. In this paper, we introduced a state-of-the-art SS-FSL model, by introducing a semi-supervised meta-training loss, namely the Prototypical Random Walk, which enforces global consistency over the data manifold, and magnetizes points around their class prototypes. We show that our model outperforms prior art and rivals its fully labeled counterpart in a wide range of experiments and analysis. We contrast the effects and performance of global versus local consistency, by training a PN with VAT~\citep{miyato2018virtual} and comparing it with our model. While the local consistency loss has an improvement on the performance, we found out that our global consistency loss significantly improves the performance in SS-FSL.
Finally, we show that our model is robust to distractor classes even when they constitute the majority of unlabeled data. We show how this is related to the dynamic of PRW. We even create a simple distractor filter, and show its efficiency in improving semi-supervised inference~\citep{ren2018metalearning}. Our experiments and results set the state-of-the-art on most benchmarks.


\bibliography{neurips.bib}

\end{document}


\onecolumn
\maketitle

This document include the following content. 
\tableofcontents

\section{miniImageNet higher way classification}\label{app:mini}
 The relative improvement over PN as we increase the number of classes as shown in Fig.~\ref{fig:tsne} on miniImageNet dataset. 
\begin{figure}[h!]
\centering
\includegraphics[width=0.45\textwidth]{mini-bars.png}\\
\includegraphics[width=0.45\textwidth]{mini_relativ.png}
\caption{\emph{Left:} Accuracy of our different models on 5-shot miniImageNet as we increase the number of classes $N_c$. \emph{Right:} The relative improvement over PN as we increase the number of classes. }
\label{fig:tsne}
\end{figure}

\section{PCA Embedding Images on Spirial and Guassian Circle datasets }\label{app:mini}
\emph{cf.}~Fig.~\ref{fig:spiral} shows the results, it can be seen that the proposed method embedding space is similar to 100\% labels embedding space hence producing decision boundary similar to 100\% labels, for both synthetic spirial and guassian circle datasets.In~Table.~\ref{table:toy}, we quantitatively show that our loss  significantly improves the performance from 89,25\% to 97.17\% close to the full data performance (99.92\%) on the Spiral tot experiment. Similarly, our loss  significantly improves the performance from 48.33\% to 79.44\%, getting much closer to the full data performance (88.89\%) on the Circle toy experiment.  
\usepackage{float}
\begin{figure*}[t]
\vspace{-1.5mm}
    \centering
    \includegraphics[width=0.90\textwidth]{aaai/figure.png}
    \caption{This is the PCA embedding plot of spiral dataset experiment and gaussian circle dataset. The colored points are unlabeled data, the "x" marks are the centroid of each labeled data. $(a)$ The toy experiments trains on synthetic spiral data set. Left: Prototypical Random Walk (PRW)  trains with 10\% labeled data, Middle: trains the base model with 10\% labeled data Right: trains the base model with Full 100\% labeled data (b) Left: Prototypical Random Walk (PRW)  trains with 5\% labeled data, Middle: trains the base model with 5\% labeled data Right: trains the base model with Full 100\% labeled data }
    \label{fig:spiral}
    \vspace{-1.5mm}
\end{figure*}

\begin{table}[h!]
\caption{2D synthetic datasets accuracy}
\label{table:toy}
\begin{tabular}{lllll}
\hline
                              & \multicolumn{2}{l}{Spiral Accuracy} & \multicolumn{2}{l}{Circle Accuracy} \\ \hline
                              & train             & val             & train                 & val                  \\ \hline
Few labeled+PRW & {\bfseries 94.67}             & {\bfseries 97.17}           & {\bfseries 80.00}                 & {\bfseries 79.44}                \\
Few labeled               & 90.67             & 89.25           & 51.00                 & 48.33                \\
Full labeled             & 100.00            & 99.92           & 86.67                 & 88.89               
\end{tabular}
\end{table}

\clearpage

\section{Episode construction}
\label{alg1}
\begin{algorithm}
   \caption{Construct a semi-supervised episode $E$, optionally with distractors. \\ 
   RANDOMSAMPLE(S, N) denotes a set of N elements chosen randomly from set $S$, without replacement.
   Items sampled from the labeled split are tuples $(x_i, y_i)$, while items sampled from the unlabeled split are simply $(x_i)$}
   \begin{algorithmic}
   \REQUIRE 
            $N_c$ \COMMENT{The number of classes or \emph{way}}\\
            $N_s$ \COMMENT{The number of examples per class or \emph{shot}}\\
            $N_q$ \COMMENT{The number of query images per class}\\
            $N_u$ \COMMENT{The number of unlabeled examples per class in the support set}\\
            $N_d$ \COMMENT{The number of distractor classes per episode}\\
            \vspace{2pt}
      \STATE  $V \leftarrow  $ RANDOMSAMPLE($\{1 \cdots K \}$ , $N_c$)
      \FOR{$k \in V$}
        \STATE $\mathcal{S}_{k,L} \leftarrow$ RANDOMSAMPLE($\mathcal{D}_{{k},L}$ , $N_s$)
        \STATE $\mathcal{S}_{k,U} \leftarrow$ RANDOMSAMPLE($\mathcal{D}_{{k},U}$ , $N_u$)
        \STATE $\mathcal{Q}_k \;\; \leftarrow$ RANDOMSAMPLE($\mathcal{D}_{{k},L}$ , $N_q$)
        \STATE $\mathcal{Q} \leftarrow \mathcal{Q}  \cup \mathcal{Q}_k$
        \STATE $\mathcal{S} \leftarrow \mathcal{S} \cup (\mathcal{S}_{k,L} \cup \mathcal{S}_{k,U}$)
      \ENDFOR
    \IF{DISTRACTOR}
        \STATE $L \leftarrow$  RANDOMSAMPLE($\{1 \cdots K \}/V$ , $N_d$)
        \FOR{$k \in L$}
            \STATE $\mathcal{S} \leftarrow \mathcal{S} \: \cup$ RANDOMSAMPLE($\mathcal{D}_{{k},U}$ , $N_u$)
        \ENDFOR
    \ENDIF
   \end{algorithmic}
\end{algorithm}

\section{Experimental details}\label{app:hyperparams}
\textbf{Architecture} We use the architecture from \cite{vinyals2016matching} composed of four
convolutional blocks. Each block comprises a 64-filter 3 × 3 convolution, batch normalization layer, a ReLU nonlinearity and a 2 $\times$ 2 max-pooling layer. No other architectures were considered

\textbf{Optimization} We use ADAM for all our experiments with $\beta_1$ = 0.9, $\beta_2$ = 0.99. The initial learning rate for Omniglot experiments 0.001, and  for all mini-imagenet 0.00025. For Omniglot, we cut the learning rate in half every 2K episodes. For mini-imagenet we cut the rate in half every 12500 episodes. For omniglot, we train for 20K episodes, for mini-imagenet we train for 100K episodes. 
For the learning rate the range considered was from 0.0001 to 0.001. For the decay steps the range considered was 2k to 20k. The maximun training step considered was 120K.

\textbf{Random walk} For all experiments the number of steps for our random walk $\tau$ = 3, with an eponential decay $\alpha$ of 0.7, except on omniglot eperiments without distractors, where $\alpha$ = 1. The weight given to for the semi-supervised loss $\lambda$ is 1.5 \& 2 respectively for omniglot without and with distractors. For mini-imagenet, $\lambda$ is 0.5 for all experiments. We validated over $\tau$ in the range from 0 to 5, $\alpha$ from 0.5 to 1, and $\lambda$ from 0.5 to 2.5.

\subsection{Episode composition}
\textbf{Omniglot} For training without distractors, we use $N_c$ = 20 and $N_u$ = 10. For training with distractors, we use $N_c = N_d$ = 5 and $N_u$ = 10. And $N_s$ = 1 in all cases. For all experiments, $N_q$ = 5.
For testing with additional unlabelled data, we use $N_u$ = 5 for fair comparison with ~\cite{ren2018metalearning}. 

\textbf{Mini-imagenet} For training we use $N_s$ = 5, $N_c$ = 5, and $N_u$ = 10 for all experiments. When training with distractors, $N_d$ = 5.  For all experiments, $N_q$ = 5.
For testing with additional unlabelled data, we use $N_u$ = 20 for fair comparison with ~\cite{ren2018metalearning}. 

\clearpage
\section{Latent space visualization}
\begin{figure}[h]
\centering
\includegraphics[width=0.45\textwidth]{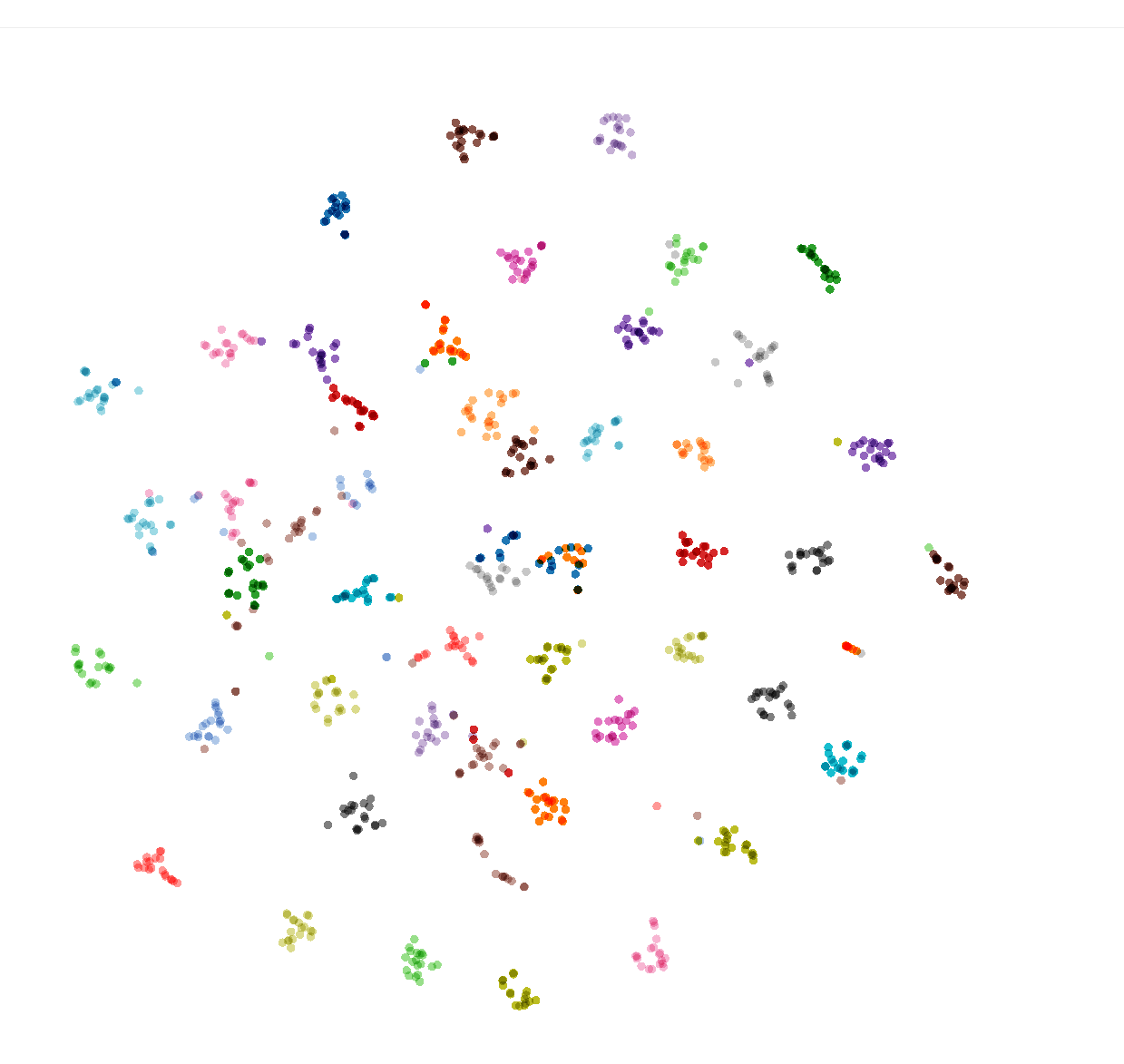}\\
\includegraphics[width=0.45\textwidth]{RW_50_19_2.png}
\caption{tSNE visualization of the embeddings for 50 classes on Omniglot. The embeddings of unlabeled data got magnetized to the class prototypes forming more compacted clusters in our PRWN \emph{right} in contrast to the embeddings of PN \emph{left}}
\label{fig:tsne}
\end{figure}

\input{aaai/deeperarch}

\section{Experiments of applying PRW loss to FEAT}
We use the group of hyper-parameters as $\tau = 1$, $\alpha=0.7$ and $N_{u}=10$ for the semi-supervised part of our experiment. Note that we compare our method with the original FEAT~\cite{ye2020few} under the same hyper-parameter settings.  Note that FEAT here is applied into our semi-supervised few-shot learning setting, where unlabeled data is used during the training. Our results, reported in Table.~\ref{tab:feat} , show that our PRW loss is indeed orthogonal to FEAT and can slightly improve the performance regardless the network architecture. 

\begin{table}[h]
    \centering
  \caption{Few-shot classification accuracy on MiniImageNet with 60\% labelled data. Note that we compare our method with the original FEAT under the same setting. For PRW loss, we use $Nu=10$, $\alpha=0.7$ and $\tau=3$}
 \scalebox{0.8}{
    \begin{tabular}{l c cc c}
         \hline
         Model&\multicolumn{2}{c}{5-way 1-shot}&\multicolumn{2}{c}{5-way 5-shot}\\
         \hline
          & Conv4 & Res12 & Conv4 & Res12
          \\
          \hline
          FEAT~\cite{ye2020few}&53.97&71.0&{\bfseries 71.31}&84.0 \\
          \hline
          FEAT~\cite{ye2020few} +PRW & {\bfseries 54.18}&{\bfseries 71.03}&71.07&{\bfseries 84.56}\\
          \hline
          
    \end{tabular}
   }
    \label{tab:feat}
 
\end{table}

\clearpage
\bibliography{neurips.bib}
\bibliographystyle{aaai}

%% file: table2.tex
\pdfoutput=1
\begin{table*}[!htbp]
\centering
    \caption{Experiments with distractor classes}\label{tbl_dist}
    \resizebox{0.9\textwidth}{!}{
    \begin{tabular}{l c c c c c} 
     \toprule
     \multirow{2}{*}{Model}             & Omniglot      &  \multicolumn{2}{ c }{Mini-Imagenet}  &  \multicolumn{2}{ c }{Tiered-Imagenet}\\ 
                                        &  1-shot       &          1-shot & 5-shot & 1-shot & 5-shot \\
    \toprule
    PRWN (Ours)                     & 97.76 $\pm$ 0.11   & 50.96 $\pm$ 0.23  & 67.64 $\pm$  0.18 & 53.30 $\pm$ 1.02 & 69.88 $\pm$ 0.96\\
    \midrule
    PN+ Semi-supervised inference~\citep{ren2018metalearning}        & 95.08 $\pm$ 0.09  & 47.42 $\pm$ 0.33   & 62.62 $\pm$ 0.24 & 48.67 $\pm$ 0.60 & 67.46 $\pm$ 0.24 \\
    PN+ Soft K-means~\citep{ren2018metalearning}                     & 95.01 $\pm$ 0.09 & 48.70 $\pm$ 0.32 & 63.55 $\pm$ 0.28 & 49.88 $\pm$ 0.52 & 68.32 $\pm$ 0.22 \\
    PN+ Soft K-means + cluster~\citep{ren2018metalearning}           & 97.17 $\pm$ 0.04 & 48.86 $\pm$ 0.32 & 61.27 $\pm$ 0.24 & 51.36 $\pm$ 0.31 & 67.56 $\pm$ 0.10\\
    PN+ Masked soft K-means~\citep{ren2018metalearning}              & 97.30 $\pm$ 0.30 & 49.04 $\pm$ 0.31 & 62.96 $\pm$ 0.14 & 51.38 $\pm$ 0.38 & 69.08 $\pm$ 0.25\\
    TPN-Semi~\citep{liu2018learning} &  N/A &  50.43 $\pm$ 0.84 & 64.95 $\pm$ 0.73 & 53.45 $\pm$ 0.93 & \textbf{69.93 $\pm$ 0.80}\\    
    PRWN+ Semi-supervised inference (Ours) & \textbf{97.86} $\pm$ \textbf{0.22}   & \textbf{53.61} $\pm$ \textbf{0.22}  & \textbf{67.45} $\pm$ \textbf{0.21} & \textbf{56.59} $\pm$ \textbf{1.13} & 69.58 $\pm$ 1.00 \\
    PRWN+ Semi-supervised inference + filter (Ours) & \textbf{99.04} $\pm$ \textbf{0.18}   & \textbf{54.51} $\pm$ \textbf{0.23}  & \textbf{68.77} $\pm$ \textbf{0.20} & \textbf{57.97} $\pm$ \textbf{1.12} & 69.74 $\pm$ 1.10\\    
    \end{tabular}
    }
\end{table*}